\documentclass[conference]{IEEEtran}
\IEEEoverridecommandlockouts
\usepackage{cite}
\usepackage{amsmath,amssymb,amsfonts}
\usepackage{algorithmic}
\usepackage{graphicx}
\usepackage{textcomp}
\usepackage{xcolor}
\usepackage{url}
\usepackage{booktabs}
\usepackage{subcaption}
\usepackage{float}
\usepackage{stfloats} 
\def\BibTeX{{\rm B\kern-.05em{\sc i\kern-.025em b}\kern-.08em
    T\kern-.1667em\lower.7ex\hbox{E}\kern-.125emX}}
\begin{document}

\title{Bayesian-Symbolic Integration for Uncertainty-Aware Parking Prediction\thanks{© 2025 IEEE. Personal use of this material is permitted. Permission from IEEE must be obtained for all other uses, including reprinting/republishing. This is the accepted version of the paper accepted at IEEE ITSC 2025. The final published version will be available via IEEE Xplore.}}

\author{
  \IEEEauthorblockN{Alireza Nezhadettehad\IEEEauthorrefmark{1}, Arkady Zaslavsky\IEEEauthorrefmark{1}, Abdur Rakib\IEEEauthorrefmark{2}, Seng W. Loke\IEEEauthorrefmark{1}}
  \IEEEauthorblockA{
    \IEEEauthorrefmark{1}School of Information Technology, Deakin University, Burwood, Australia\\
    Email: \{anezhadettehad, arkady.zaslavsky, seng.loke\}@deakin.edu.au\\\
    \IEEEauthorrefmark{2}Centre for Future Transport and Cities, Coventry University, Coventry CV1 5FB, UK\\
    Email: ad9812@coventry.ac.uk\
  }
}


\maketitle

\begin{abstract}
Accurate parking availability prediction is critical for intelligent transportation systems, but real-world deployments often face data sparsity, noise, and unpredictable changes. Addressing these challenges requires models that are not only accurate but also uncertainty-aware. In this work, we propose a loosely coupled neuro-symbolic framework that integrates Bayesian Neural Networks (BNNs) with symbolic reasoning to enhance robustness in uncertain environments. BNNs quantify predictive uncertainty, while symbolic knowledge—extracted via decision trees and encoded using probabilistic logic programming—is leveraged in two hybrid strategies: (1) using symbolic reasoning as a fallback when BNN confidence is low, and (2) refining output classes based on symbolic constraints before reapplying the BNN. We evaluate both strategies on real-world parking data under full, sparse, and noisy conditions. Results demonstrate that both hybrid methods outperform symbolic reasoning alone, and the context-refinement strategy consistently exceeds the performance of Long Short-Term Memory (LSTM) networks and BNN baselines across all prediction windows. Our findings highlight the potential of modular neuro-symbolic integration in real-world, uncertainty-prone prediction tasks.
\end{abstract}

\begin{IEEEkeywords}
Neuro-symbolic integration, Bayesian neural networks, Uncertainty-aware prediction, Parking availability, Probabilistic logic programming, Smart transportation, Data scarcity, Symbolic reasoning
\end{IEEEkeywords}

\section{Introduction}
\label{sec:introduction}

Parking management has become an increasingly critical issue in the context of urban mobility, driven by growing population density and rising vehicle ownership. One of the most persistent challenges is the time wasted by drivers searching for available parking, which contributes to congestion, excessive fuel consumption, and increased carbon emissions \cite{Atif2020}. A recent study by an online parking platform reports that Australians spend an estimated 3,120 hours over their lifetime—equivalent to 130 days or over four months—searching for parking. Survey results further revealed that over 43\% of drivers typically spend more than 15 minutes per trip trying to find a space, with 12\% indicating their search often exceeds 20 minutes \cite{battaglia_australians_2022}. In the United States, similar trends have been observed, where drivers lose an average of 17 hours per year searching for parking, resulting in an annual cost of approximately \$345 per person in wasted time, fuel, and emissions \cite{mccoy_drivers_nodate}. This problem is further amplified in densely populated cities; for instance, in New York City, drivers spend an average of 107 hours annually looking for parking, incurring costs of up to \$2,243 per driver. Traditional parking prediction methods often lack the flexibility to adapt to the highly dynamic and context-sensitive nature of urban environments, where availability frequently varies with factors such as time of day, weather conditions, and local events \cite{Yang2019, Balmer2021}.

The advent of Internet of Things (IoT) technologies has enabled the deployment of smart parking systems capable of collecting real-time occupancy data via distributed sensors. This data, when combined with advanced machine learning techniques, enables predictive analytics that enhance operational efficiency and support context-aware decision-making \cite{Fahim2021, Ke2021}. For instance, Ali et al.\ proposed a smart parking system powered by a deep Long Short-Term Memory (LSTM) network that effectively captures temporal patterns in parking behavior to predict availability \cite{Ali2020}. These innovations not only benefit drivers through real-time guidance but also aid urban planners in optimizing infrastructure and policy.

However, parking prediction systems often operate under challenging conditions, including noisy or incomplete data, limited ground truth, and rapidly changing patterns. To address these limitations, recent research has explored hybrid frameworks that incorporate deep learning with representation learning techniques, enabling the integration of heterogeneous data sources such as historical logs, environmental features, and user interactions \cite{Yang2022, Zhang2022}. While these approaches offer improved predictive performance and flexibility, they often fall short in effectively managing predictive uncertainty, offering transparent decision-making, and supporting symbolic reasoning—capabilities that are especially critical in public-facing urban systems where interpretability and robustness are essential. To bridge this gap, we propose a hybrid framework that combines Bayesian Neural Networks (BNNs) for probabilistic modelling with Neuro-Symbolic AI (NS-AI) to provide structured, interpretable, and context-aware reasoning under uncertainty.

Recent advances in BNNs and Neuro-Symbolic AI hold promise for further improving these systems. BNNs offer principled mechanisms for quantifying uncertainty, which is crucial for applications that involve partial observability and noisy signals \cite{Jospin2022, Zeng2022}. On the other hand, Neuro-Symbolic AI integrates neural learning with symbolic reasoning to produce interpretable and logic-consistent predictions. This fusion provides both predictive power and transparency—essential for public-facing urban systems \cite{Garcez2023, Sarker2021}.

This paper proposes a novel loosely coupled neuro-symbolic framework for uncertainty-aware parking prediction. By integrating Bayesian neural inference with symbolic reasoning modules, the proposed approach provides both accurate forecasts and interpretable fallback mechanisms under high-uncertainty conditions. Through a series of experiments on real-world parking datasets, we demonstrate the advantages of this hybrid model under full-data, sparse-data, and noisy-data settings.

The remainder of this paper is organised as follows. Section~\ref{sec:relatedwork} reviews relevant literature on parking prediction, Bayesian Neural Networks, and Neuro-Symbolic AI. Section~\ref{sec:methodology} presents the proposed hybrid neuro-symbolic framework, including the dataset, model architecture, integration strategies, and experimental setup. Section~\ref{sec:results_and_discussion} reports and analyses the performance of the proposed models under baseline, data scarcity, and noisy input scenarios. Finally, Section~\ref{sec:conclusion_future_work} concludes the paper and outlines directions for future work.

\section{Literature Review}
\label{sec:relatedwork}

Urban parking prediction has evolved significantly with the advancement of machine learning and intelligent sensing technologies. Early approaches in parking prediction predominantly focused on static models or rule-based heuristics, but these methods quickly proved inadequate in capturing the temporal and contextual complexity of real-world urban environments. As a result, deep learning methods—particularly recurrent neural networks like Long Short-Term Memory (LSTM) and Gated Recurrent Units (GRUs)—have become popular due to their ability to model temporal dependencies in sequential data. Zeng et al.\ developed a Stacked GRU-LSTM framework that incorporated multi-factor influences such as time, location, and environmental attributes, demonstrating enhanced accuracy for real-time occupancy prediction \cite{Zeng2022}. Similarly, Shao et al.\ showed that LSTM models can effectively represent complex occupancy trends based on historical parking data, making them a viable foundation for adaptive parking systems \cite{Shao2019}.

Beyond temporal learning, contextual models have integrated additional spatial, temporal, and event-based information to capture the multifaceted nature of parking demand. For instance, Jelen et al.\ and Balmer et al.\ leveraged geographic and temporal signals to develop contextual predictors capable of adapting to localised urban variations \cite{Jelen2021, Balmer2021}. These findings highlight the importance of location-sensitive features in improving predictive generalisability. Comparative studies, such as the one by Awan et al., have also emphasised that simpler models like decision trees can remain competitive under specific configurations, though deep learning methods generally offer superior scalability and feature learning \cite{Awan2020}.

The integration of Internet of Things (IoT) infrastructure has further enriched the data landscape for parking systems. Real-time sensor deployments enable continuous monitoring of parking space availability and user mobility. Atif et al.\ and Fahim et al.\ surveyed smart parking systems that harness IoT-enabled devices and wireless sensor networks to collect real-time data and feed it into cloud-based analytics engines \cite{Atif2020, Fahim2021}. These frameworks have formed the foundation for smart city initiatives aiming to reduce congestion and environmental impact. Building on this, Ali et al.\ employed deep LSTM networks to analyse IoT sensor data from urban environments, achieving high spatial and temporal resolution in parking predictions \cite{Ali2020}. Complementary work by Badii et al.\ demonstrated the value of open data sources such as traffic feeds and weather logs to augment predictive inputs, while Bock et al.\ explored participatory sensing through a crowd of taxis acting as mobile probes for on-street space availability \cite{Badii2018, Bock2020}.

While deep learning models have achieved strong predictive performance, they often lack the ability to quantify uncertainty—an essential feature in high-variability urban environments. Bayesian Neural Networks (BNNs) offer a solution by modelling uncertainty over weights, producing probabilistic outputs rather than point estimates. Jospin et al.\ provided a tutorial on the implementation and advantages of BNNs, highlighting their applicability to safety-critical domains \cite{Jospin2022}. Zeng et al.\ applied a Bayesian approach to parking prediction, illustrating that BNNs can not only provide reliable predictions but also quantify uncertainty, a critical feature when data is missing or noisy \cite{Zeng2022}. The versatility of BNNs has also been illustrated by Jang and Lee in the context of financial forecasting, suggesting that such techniques generalise well to complex time-series applications like urban parking \cite{Jang2017}.

To address the interpretability limitations of neural models, researchers have increasingly turned to Neuro-Symbolic AI—a hybrid paradigm that combines neural network learning with logical reasoning. Garcez and Lamb formalised a taxonomy of integration strategies ranging from loosely coupled to fully differentiable neuro-symbolic systems, emphasising their role in producing explainable and robust models \cite{Garcez2023}. These models are particularly attractive in public infrastructure contexts where decisions must be transparent. Susskind et al.\ discussed practical implementations of neuro-symbolic AI, highlighting its relevance for structured reasoning and policy-driven applications \cite{Susskind2021}. Further studies by Sarker et al.\ and Ebrahimi et al.\ reviewed the application of neuro-symbolic frameworks in urban systems and knowledge representation tasks, noting their ability to fuse high-dimensional data with symbolic constraints for interpretable learning \cite{Sarker2021, Ebrahimi2021}. Oltramari et al.\ demonstrated the use of such models in context-aware environments, while Sen et al.\ developed rule-extraction methods that remain robust under noisy supervision—both being directly relevant to urban parking domains where data quality is often suboptimal \cite{Oltramari2020, Sen2022}.

Together, these developments form a strong foundation for integrating probabilistic and symbolic reasoning in parking prediction. While several neuro-symbolic approaches have emerged in recent years—typically combining conventional deep neural networks with symbolic logic—they often rely on deterministic models and do not explicitly account for uncertainty. To the best of our knowledge, no existing work has incorporated Bayesian Neural Networks (BNNs) as the neural backbone of a neuro-symbolic system for this task. This paper addresses that gap by introducing a hybrid architecture that combines the uncertainty modelling capabilities of BNNs with the interpretability of symbolic reasoning. The proposed method is designed to perform reliably in challenging real-world scenarios characterised by data scarcity, epistemic uncertainty, and noisy observations.

\section{Methodology}
\label{sec:methodology}

This section outlines the proposed methodology for uncertainty-aware parking occupancy prediction using a hybrid neuro-symbolic architecture. The approach combines Bayesian Neural Networks (BNNs) with rule-based reasoning, enabling the system to make accurate and interpretable predictions even in the presence of uncertainty, noise, or limited training data.

\subsection{Utilised Dataset and Preprocessing}

The dataset used in this study is sourced from Melbourne’s open data portal \cite{melbourneparking}, which provides real-time occupancy information collected from on-street parking bay sensors installed across the city's central business district (CBD). These in-ground sensors detect vehicle arrivals and departures and generate event-based records that include the parking bay ID, timestamp, sensor status (occupied or unoccupied), stay duration, geospatial coordinates, applicable parking restriction, and whether the vehicle overstayed the permitted limit. The full dataset contains approximately 42.7 million records for the year of 2019.

Since raw entries can include inconsistencies due to communication failures or sensor malfunctions, a data-cleaning process was first applied to ensure high integrity. The cleaned data was then aggregated into fixed 15-minute time intervals at the street segment level, where each segment refers to the group of parking bays located between two adjacent road junctions. For each time slot, the occupancy ratio was computed as the proportion of occupied bays within a segment. 

To capture temporal dynamics in parking behaviour, the model uses both current and past occupancy ratios as input features. These temporal inputs are combined with a variety of contextual features, including time of day, day of week, month, public holiday indicator, and environmental conditions such as weather type, temperature, wind speed, and rainfall. These features help the model account for both regular usage patterns and irregular fluctuations in parking demand.

The dataset covers all street segments in Melbourne's CBD, and the BNN is trained using the full spatiotemporal history of these segments. While we do not explicitly model spatial correlations between neighbouring segments, the use of aggregated data at the segment level ensures that each prediction is contextually grounded in its local environment. The model thus implicitly captures spatial variability across different parts of the CBD.

The target variable—the parking occupancy ratio—was discretised into five equally sized classes: \textit{Very Low}, \textit{Low}, \textit{Moderate}, \textit{High}, and \textit{Very High}, each corresponding to a 20\% occupancy range. This classification-based formulation allows the model to produce structured outputs and to explicitly represent predictive uncertainty.

\subsection{Bayesian Neural Network with Confidence Thresholding}
To model uncertainty in prediction, we adopt a Bayesian Neural Network trained using variational inference. The model places a posterior distribution over its weights and biases, and represents the output as a probability distribution over possible outcomes. This is achieved by integrating over the posterior distribution of the model parameters:
\[
p(y \mid \mathbf{x}, \mathcal{D}) = \int p(y \mid \mathbf{x}, \boldsymbol{\theta}) \, p(\boldsymbol{\theta} \mid \mathcal{D}) \, d\boldsymbol{\theta}
\]
where $\mathbf{x}$ is the input, $y$ is the output, $\boldsymbol{\theta}$ denotes the weights and biases of the network, $p(y \mid \mathbf{x}, \boldsymbol{\theta})$ is the likelihood function, and $p(\boldsymbol{\theta} \mid \mathcal{D})$ is the posterior distribution over the parameters given the training data $\mathcal{D}$.

To make predictions for a new input $\mathbf{x}^*$, the model uses the posterior predictive distribution:
\[
p(y^* \mid \mathbf{x}^*, \mathcal{D}) = \int p(y^* \mid \mathbf{x}^*, \boldsymbol{\theta}) \, p(\boldsymbol{\theta} \mid \mathcal{D}) \, d\boldsymbol{\theta}
\]
where $\mathbf{x}^*$ is a new input sample and $y^*$ is the corresponding predicted output.

To improve decision quality, particularly under uncertain conditions, we introduce a confidence thresholding mechanism. Specifically, a prediction is considered \textit{valid} only if the highest class probability exceeds a predefined confidence threshold. We experimented with several threshold values and found that a 30\% threshold provides the best trade-off between prediction ratio (the proportion of test instances for which a confident prediction is made) and accuracy. When this threshold is not met, the model is considered uncertain about its prediction, and the task is handed off to a symbolic reasoning module for assistance. This approach ensures that only well-calibrated probabilistic decisions are accepted from the BNN, while uncertain cases are handled through interpretable, rule-based logic.

\subsection{Hybrid Neuro-Symbolic Framework}
\begin{figure}[t]
    \centering
    \includegraphics[width=0.9\linewidth]{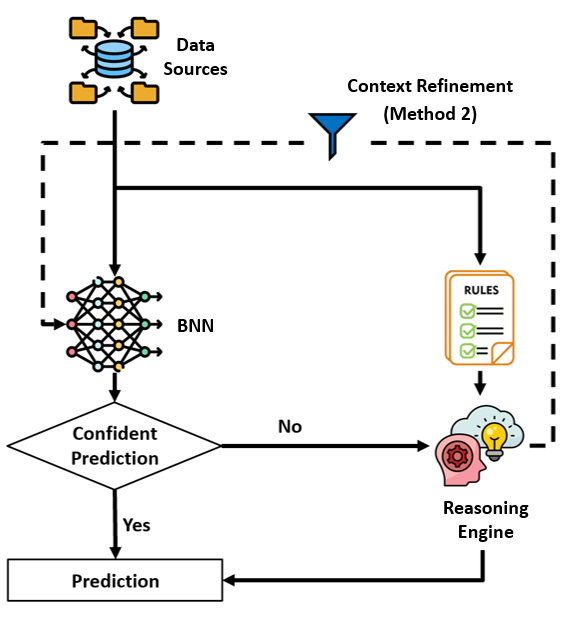}
    \caption{Overview of the hybrid neuro-symbolic decision process. Method 1 uses symbolic reasoning as a fallback when BNN confidence is low. Method 2 applies symbolic constraints to refine the output space before re-evaluating BNN predictions.}
    \label{fig:methodology}
\end{figure}

The key contribution of this study is a hybrid framework that combines neural and symbolic reasoning. The system first uses the BNN with 30\% confidence thresholding for selective prediction to make a prediction and estimate its confidence. If the BNN is confident, its prediction is accepted. Otherwise, one of two symbolic integration strategies is invoked (see Fig.~\ref{fig:methodology}):

\begin{itemize}
    \item \textbf{Method 1 (Fallback to Reasoning):} If the BNN's confidence is below the 30\% threshold, the symbolic reasoning module is directly used to generate a prediction based on rule-based inference.
    
    \item \textbf{Method 2 (Contextual Refinement):} In this more sophisticated strategy, the symbolic module restricts the BNN's output space by eliminating implausible classes. The BNN then re-evaluates its prediction over this refined set. If the confidence within the refined space exceeds the threshold, the BNN's updated prediction is accepted. If not, the symbolic reasoning output is used.
\end{itemize}

Symbolic rules are extracted from the training data using decision tree induction and encoded into a Probabilistic Logic Program (PLP). These rules capture interpretable patterns—such as time-of-day effects, weekday/weekend distinctions, and environmental influences—that complement the BNN’s capacity to model complex, non-linear relationships. By expressing high-confidence patterns as logic-based rules, the symbolic module provides an interpretable fallback or refinement mechanism when the BNN is uncertain.

Examples of typical extracted rules include:

\begin{align*}
& \text{If } \texttt{Hour} \in [12,14] \text{ and } \texttt{Day} \in \{\text{Mon--Fri}\} \\
& \quad \text{and } \texttt{PrevOcc} = \text{High} \Rightarrow \texttt{Occ} = \text{VeryHigh} \\
\\
& \text{If } \texttt{Day} \in \{\text{Sat, Sun}\} \text{ and } \texttt{Hour} \in [9,11] \\
& \quad \text{and } \texttt{PrevOcc} = \text{Low} \Rightarrow \texttt{Occ} = \text{Moderate} \\
\\
& \text{If } \texttt{Rainfall} > 1.0 \text{ and } \texttt{PrevOcc} = \text{VeryHigh} \\
& \quad \Rightarrow \texttt{Occ} = \text{High}
\end{align*}

These rules are used by the symbolic module to either produce predictions directly (in fallback mode) or constrain the neural output space (in refinement mode).

In Method 2, the symbolic module helps refine the BNN's output space by eliminating implausible classes based on the contextual rules. For example, if a PLP (based on rules extracted by decision tree) indicates that \textit{Very High} occupancy is unlikely on weekends before 9 AM, the class \texttt{VeryHigh} will be excluded from the set of candidate outputs in such contexts. The BNN then re-evaluates its prediction using only the remaining plausible classes—e.g., \{\texttt{Low}, \texttt{Moderate}, \texttt{High}\}—and checks whether any of them surpass the confidence threshold. If so, the refined prediction is accepted. Otherwise, the final prediction defaults to the symbolic output.

This process ensures that predictions made under uncertainty are not only guided by learned distributions but are also consistent with interpretable domain rules. It acts as a soft constraint mechanism, narrowing the decision space in ways that reflect structural regularities observed in the data.

\subsection{Handling Data Scarcity and Noise}

Real-world parking data is often incomplete or noisy. To assess model robustness, we simulate both data scarcity and input noise. For data scarcity, we train models on progressively smaller subsets of the dataset—90\%, 50\%, and 10\%—which are randomly sampled from the full training data to evaluate how performance degrades as data becomes limited.
For noise, we inject synthetic noise into both the feature set and class labels, mimicking sensor faults and environmental anomalies.

Under these conditions, the symbolic module plays a particularly important role. When the BNN is uncertain due to sparse or noisy input, rule-based reasoning steps in to preserve prediction reliability.

\subsection{Evaluation Metrics}

We evaluate model performance using two complementary metrics: \textbf{Accuracy} and \textbf{Accuracy@1}. 

Standard accuracy measures the proportion of exact class matches between the model's predictions and the true occupancy classes.

However, given the inherent variability in occupancy data—and the ordinal nature of the class labels—we also report Accuracy@1. This relaxed metric considers a prediction correct if it matches either the true class or a neighbouring class (i.e., one bin away). For example, if the true class is \textit{High} and the model predicts \textit{Moderate} or \textit{Very High}, it is still considered an acceptable prediction under Accuracy@1. This accounts for minor deviations that would be practically tolerable in real-world scenarios.

\section{Results and Discussion}
\label{sec:results_and_discussion}

This section presents the evaluation of our proposed neuro-symbolic prediction framework, comparing its performance against three baseline models: a traditional Long Short-Term Memory (LSTM) network, a Bayesian Neural Network (BNN), and a symbolic reasoning system based on probabilistic logic. 

\subsection{Experimental Setup}
The experiments were conducted on a cluster of NVIDIA GPUs using TensorFlow and TensorFlow Probability (TFP). The Melbourne city parking dataset \cite{melbourneparking} was used, with a training-validation-test split of 80\%-10\%-10\%, ensuring temporal consistency. Models were trained using the Adam optimizer with an initial learning rate of 0.001, batch size of 64, and early stopping to prevent overfitting.

The LSTM model consists of three layers with 1500, 1500, and 1000 hidden units and serves as a strong temporal sequence learning baseline without uncertainty modelling. The BNN, trained via variational inference, incorporates uncertainty estimation and applies a 30\% confidence threshold to filter out low-confidence predictions. The symbolic reasoning system is built using rules extracted via decision tree induction and encoded into a Probabilistic Logic Program (PLP), offering interpretable and structured predictions.

Our primary objective is to evaluate the effectiveness of two proposed neuro-symbolic hybrid methods that integrate BNN predictions with symbolic reasoning. The five evaluated models are:
\begin{itemize}
    \item \textbf{LSTM (Baseline)}: A three-layer LSTM with hidden sizes [1500, 1500, 1000].
    \item \textbf{BNN (Baseline)}: A Bayesian Neural Network with the same number of layers and neurons as LSTM.
    \item \textbf{Symbolic Reasoning}: A purely rule-based model relying on Probabilistic Logic Programming (PLP), using decision tree-extracted rules to make predictions.
    \item \textbf{BNN + Symbolic (Method 1)}: A hybrid model where predictions falling below the BNN's confidence threshold are delegated to the symbolic reasoning engine.
    \item \textbf{BNN + Symbolic (Method 2)}: An advanced integration where symbolic reasoning is used to constrain the BNN's output space during low-confidence predictions. If the BNN remains uncertain, the symbolic output is used as the final prediction.
\end{itemize}

To assess performance, all models are tested across three experimental conditions—full data, data scarcity, and noisy inputs—and evaluated using two metrics: Accuracy and Accuracy@1. Each metric is reported over three consecutive 15-minute prediction windows (PW1, PW2, PW3). A 15-minute horizon is selected to reflect the operational needs of real-time parking systems, where timely and short-term forecasts are crucial for enabling dynamic decision-making. Specifically, PW1 corresponds to predictions for the 0–15 minute interval, PW2 for 15–30 minutes ahead, and PW3 for 30–45 minutes. Each window is treated independently to assess model performance across multiple near-future intervals. Tables~\ref{tab:baseline_results}--\ref{tab:noise_results} summarise the comparative results.

\subsection{Baseline}

Table~\ref{tab:baseline_results} presents the results under full data conditions. While LSTM achieves the highest raw Accuracy among the baseline models, and BNN demonstrates improved calibration reflected in higher Accuracy@1, both exhibit limitations in handling uncertainty. The symbolic reasoning model, though interpretable, underperforms due to its limited capacity to model complex patterns in the data.

In contrast, the neuro-symbolic methods demonstrate significant gains, particularly Method 2. By incorporating symbolic constraints into the BNN’s decision process, Method 2 achieves the highest performance across all prediction windows, surpassing both neural and symbolic baselines in both Accuracy and Accuracy@1. This confirms that symbolic refinement not only compensates for low-confidence predictions but also improves the overall quality of BNN outputs by narrowing the decision space in uncertain cases.

To better understand the role of symbolic reasoning in the hybrid models, we analysed how often the BNN defers prediction due to low confidence in our prior work ~\cite{nezhadettehad2025uncertainty}. In the baseline (full data) scenario, approximately 40\% of test instances were flagged as uncertain by the BNN and thus required symbolic intervention. Under data scarcity conditions, this rate increased slightly, ranging from 46\% to 52\% across the 90\%, 50\%, and 10\% training splits. In the noisy input scenario, about 47\% of the predictions fell below the confidence threshold and were delegated to the symbolic module. These results highlight the critical role of symbolic reasoning in maintaining model reliability under uncertainty, and validate the importance of designing hybrid mechanisms that can effectively handle such fallback cases. For a detailed discussion of confidence-based deferral in the context of our experiments, we refer the reader to our prior work.

\begin{figure}[!t]
    \centering
    \includegraphics[width=\linewidth]{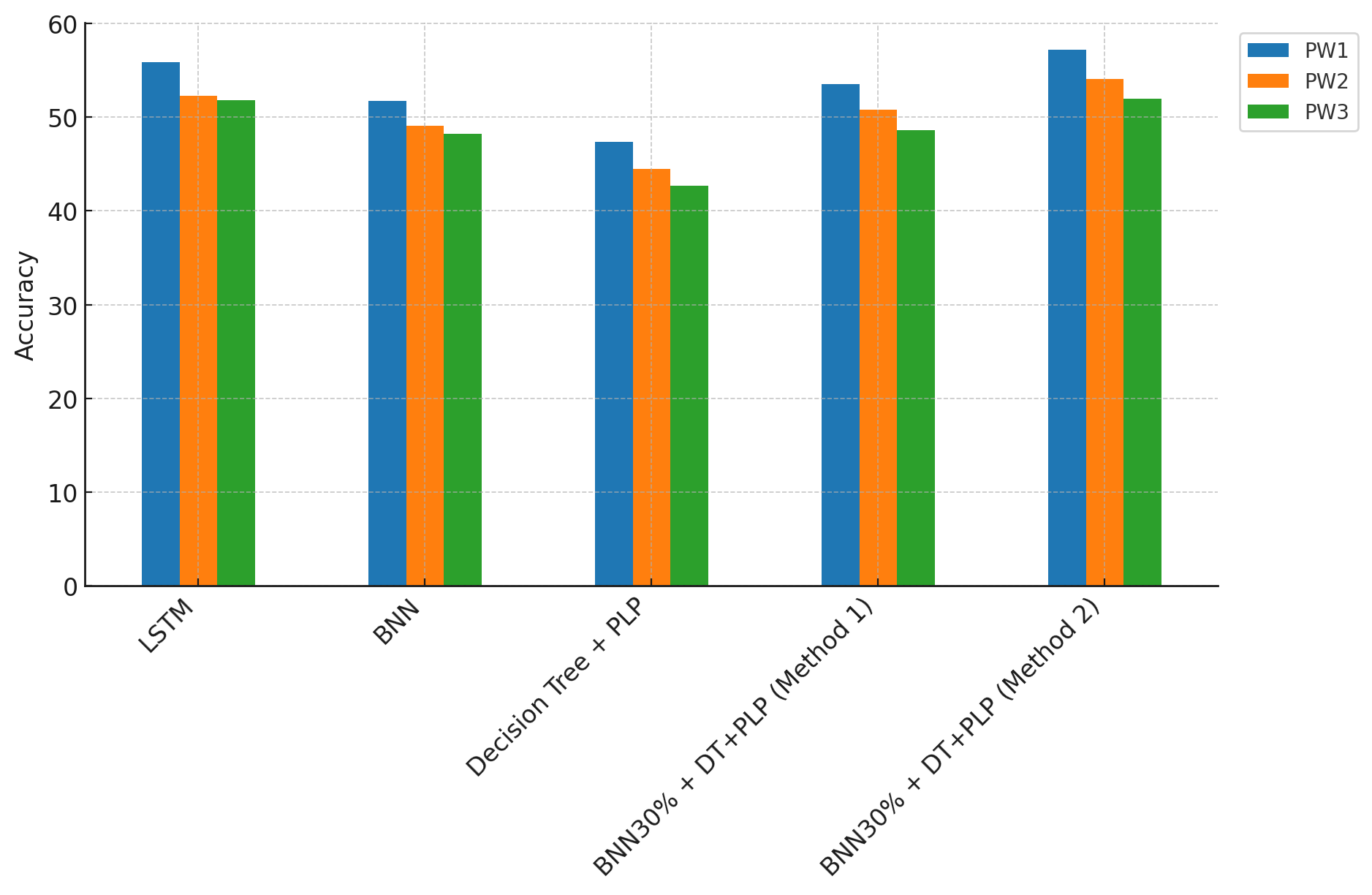}
    \caption{Accuracy under full data (baseline) condition.}
    \label{fig:baseline_acc}
\end{figure}

Method 1, while more lightweight, still consistently outperforms the standalone symbolic model and provides a robust fallback mechanism when the BNN is uncertain. Figure~\ref{fig:baseline_acc} further illustrate these trends, where the neuro-symbolic methods—especially Method 2—show clear advantages in both exact and relaxed classification accuracy across all prediction windows.

\begin{table*}
\centering
\caption{Performance of All Models Under Full Data (100\%) Condition}
\label{tab:baseline_results}
\begin{tabular}{lcccccc}
\toprule
                     Model &  PW1 Acc &  PW1 Acc@1 &  PW2 Acc &  PW2 Acc@1 &  PW3 Acc &  PW3 Acc@1 \\
\midrule
                      LSTM &     55.9 &       90.2 &     52.3 &       89.1 &     51.8 &       88.0 \\
                       BNN &     51.7 &       89.2 &     49.1 &       87.5 &     48.2 &       86.3 \\
       Decision Tree + PLP &     47.4 &       81.3 &     44.5 &       78.1 &     42.7 &       76.3 \\
BNN30\% + DT+PLP (Method 1) &     53.5 &       88.6 &     50.8 &       86.2 &     48.6 &       85.2 \\
BNN30\% + DT+PLP (Method 2) & \textbf{57.2} & \textbf{91.0} & \textbf{54.1} & \textbf{89.4} & \textbf{52.0} & \textbf{88.5} \\
\bottomrule
\end{tabular}
\end{table*}

\subsection{Impact of Data Scarcity}

Table~\ref{tab:data_scarcity_results} reports the performance of all models when training data is progressively reduced to 90\%, 50\%, and 10\% of the full dataset. As expected, both LSTM and BNN experience substantial declines in Accuracy and Accuracy@1 as data becomes limited, reflecting their reliance on sufficient training examples to generalise effectively. The symbolic reasoning model, while overall less accurate, shows relatively modest performance degradation, owing to its dependence on structural rules rather than data-driven learning.

The neuro-symbolic models demonstrate superior resilience in these low-data regimes. Method 1 consistently outperforms the symbolic-only approach across all scarcity levels, validating the benefit of selectively delegating low-confidence cases to a logic-based reasoning engine. More importantly, Method 2 maintains leading performance in both Accuracy and Accuracy@1 even when trained with only 10\% of the original data. This highlights the advantage of using symbolic constraints to refine the prediction space, allowing the BNN to make more informed decisions even under severely limited supervision.

\begin{figure*}[!t]
    \centering
    \includegraphics[width=\textwidth]{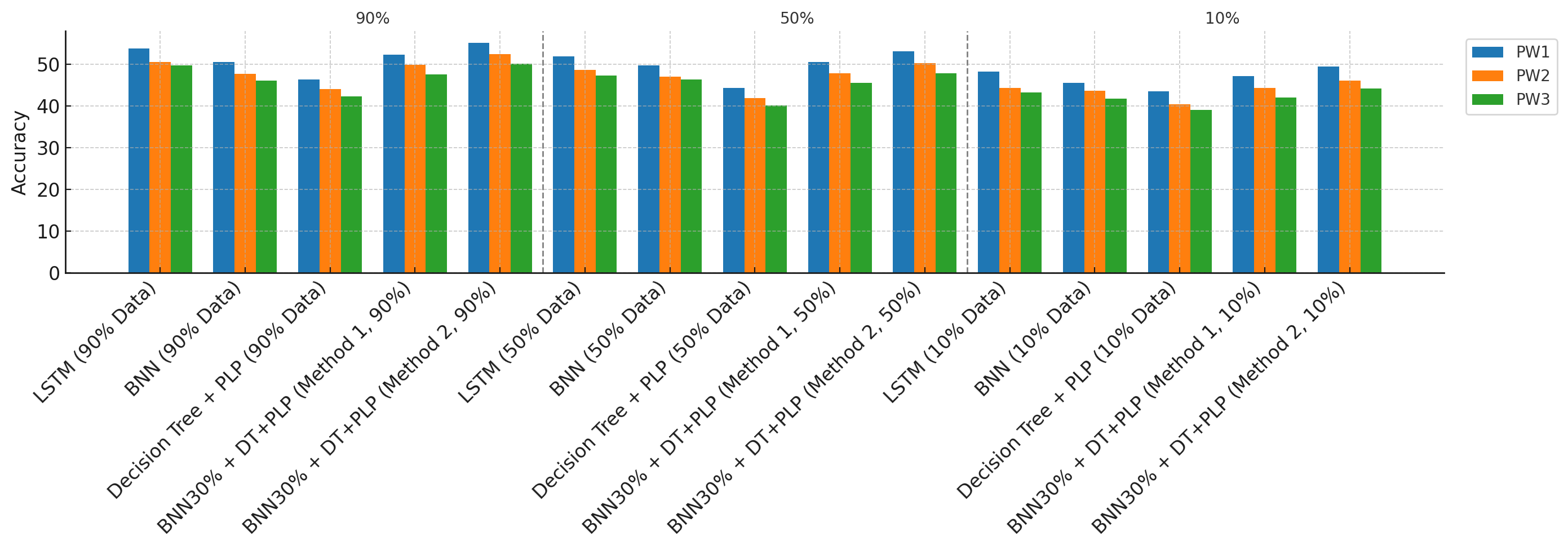}
    \caption{Accuracy under data scarcity conditions (90\%, 50\%, and 10\%).}
    \label{fig:scarcity_acc}
\end{figure*}

Figure~\ref{fig:scarcity_acc} further illustrates these trends across the three prediction windows. The hybrid models—particularly Method 2—not only slow down the degradation observed in LSTM and BNN but also establish new performance upper bounds under scarcity, underscoring the value of hybrid reasoning in data-constrained environments.

\begin{table*}
\centering
\caption{Performance of All Models Under Data Scarcity Conditions (90\%, 50\%, and 10\%)}
\label{tab:data_scarcity_results}
\begin{tabular}{lcccccc}
\toprule
                          Model &  PW1\_Acc &  PW1\_Acc@1 &  PW2\_Acc &  PW2\_Acc@1 &  PW3\_Acc &  PW3\_Acc@1 \\
\midrule
                LSTM (90\% Data) &    53.80 &      88.40 &    50.60 &      86.90 &    49.80 &      85.70 \\
                 BNN (90\% Data) &    50.50 &      88.30 &    47.70 &      86.70 &    46.10 &      85.40 \\
 Decision Tree + PLP (90\% Data) &    46.40 &      83.60 &    44.10 &      80.90 &    42.30 &      78.50 \\
BNN30\% + DT+PLP (Method 1, 90\%) &    52.30 &      87.80 &    49.90 &      85.80 &    47.60 &      84.00 \\
BNN30\% + DT+PLP (Method 2, 90\%) &    \textbf{55.20} &      \textbf{89.50} &    \textbf{52.40} &      \textbf{88.10} &    \textbf{50.20} &      \textbf{86.70} \\
\hline
                LSTM (50\% Data) &    51.90 &      84.80 &    48.70 &      83.60 &    47.30 &      82.70 \\
                 BNN (50\% Data) &    49.70 &      \textbf{86.90} &    47.10 &      \textbf{85.20} &    46.30 &      84.20 \\
 Decision Tree + PLP (50\% Data) &    44.40 &      80.70 &    41.90 &      77.90 &    40.10 &      75.40 \\
BNN30\% + DT+PLP (Method 1, 50\%) &    50.60 &      83.50 &    47.80 &      81.20 &    45.50 &      79.60 \\
BNN30\% + DT+PLP (Method 2, 50\%) &    \textbf{53.10} &      86.20 &    \textbf{50.30} &      84.40 &    \textbf{47.90} &      \textbf{82.30} \\
\hline
                LSTM (10\% Data) &    48.30 &      79.60 &    44.40 &      77.30 &    43.20 &      76.20 \\
                 BNN (10\% Data) &    45.60 &      82.70 &    43.70 &      80.60 &    41.80 &      78.80 \\
 Decision Tree + PLP (10\% Data) &    43.56 &      75.81 &    40.44 &      73.40 &    39.08 &      71.78 \\
BNN30\% + DT+PLP (Method 1, 10\%) &    47.20 &      81.50 &    44.30 &      79.00 &    42.00 &      77.30 \\
BNN30\% + DT+PLP (Method 2, 10\%) &    \textbf{49.50} &      \textbf{83.60} &    \textbf{46.10} &      \textbf{81.40} &    \textbf{44.20} &      \textbf{79.50} \\
\bottomrule
\end{tabular}
\end{table*}

\subsection{Impact of Noise on Model Robustness}

Table~\ref{tab:noise_results} summarises model performance under noisy conditions, where artificial perturbations were applied to both feature inputs and class labels to simulate real-world sensor and annotation errors. As anticipated, all models experience performance degradation. The LSTM and BNN models are particularly affected, exhibiting notable drops in both Accuracy and Accuracy@1 due to their sensitivity to data-level inconsistencies. Despite its overall lower baseline performance, the symbolic reasoning model proves to be relatively robust, as its rule-based inference mechanism is less influenced by local input perturbations.

Among all methods, the neuro-symbolic models demonstrate the greatest robustness to noise. Method 1 consistently outperforms symbolic reasoning alone, confirming that even a simple uncertainty-guided fallback strategy enhances resilience in the face of corrupted data. Method 2 again achieves the highest performance across all prediction windows, maintaining strong Accuracy and Accuracy@1 scores despite the presence of noise. This indicates that symbolic context refinement not only helps disambiguate uncertain neural predictions but also absorbs noise by constraining the output space based on stable, interpretable rules.

\begin{figure}[!t]
    \centering
    \includegraphics[width=\linewidth]{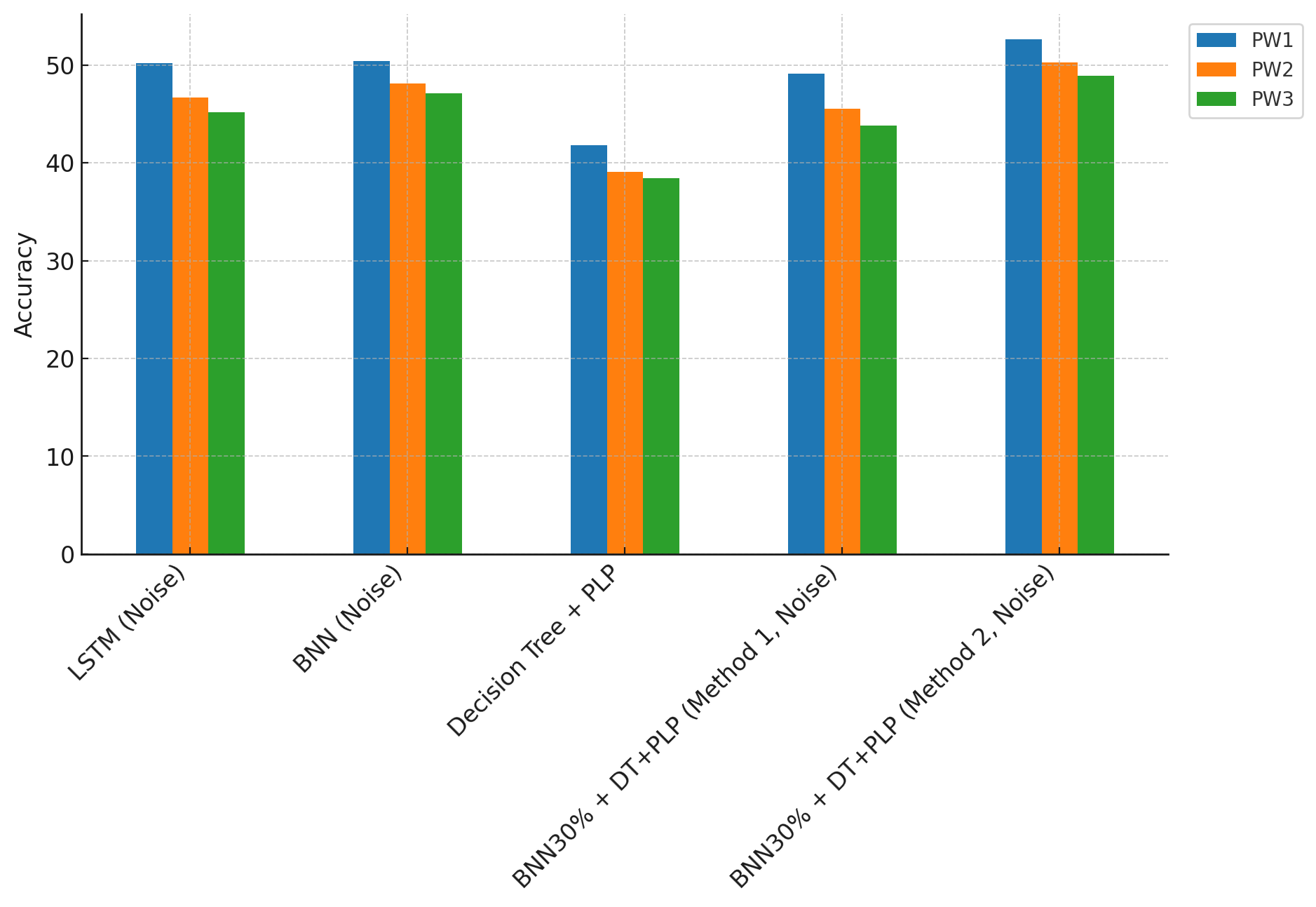}
    \caption{Accuracy under noisy input conditions.}
    \label{fig:noise_acc}
\end{figure}

Figure~\ref{fig:noise_acc} visually reinforces these findings. While performance declines are visible across the board, Method 2 exhibits a notably flatter trajectory, retaining a consistent advantage over all other models. This highlights the hybrid framework’s ability to buffer against aleatoric uncertainty, making it especially well-suited for deployment in noisy, real-world environments.

\begin{table*}
\centering
\caption{Performance of All Models Under Noisy Input Conditions}
\label{tab:noise_results}
\begin{tabular}{lcccccc}
\toprule
                            Model &  PW1 Acc &  PW1 Acc@1 &  PW2 Acc &  PW2 Acc@1 &  PW3 Acc &  PW3 Acc@1 \\
\midrule
                     LSTM (Noise) &     50.2 &       84.6 &     46.7 &       82.9 &     45.2 &       81.6 \\
                      BNN (Noise) &     50.4 &       88.2 &     48.1 &       86.7 &     47.1 &       85.4 \\
              Decision Tree + PLP &     41.8 &       77.5 &     39.1 &       74.3 &     38.4 &       73.0 \\
BNN30\% + DT+PLP (Method 1, Noise) &     49.1 &       83.4 &     45.5 &       81.8 &     43.8 &       80.4 \\
BNN30\% + DT+PLP (Method 2, Noise) & \textbf{52.6} & \textbf{89.1} & \textbf{50.3} & \textbf{87.5} & \textbf{48.9} & \textbf{86.2} \\
\bottomrule
\end{tabular}
\end{table*}

Across all scenarios, the symbolic refinement strategy (Method 2) consistently delivers the highest performance, particularly in conditions of data scarcity and noise. This suggests that even loosely coupled neuro-symbolic integration can enhance robustness and generalisation in real-world prediction tasks. The fallback method (Method 1) also proves effective, offering a lightweight mechanism to defer uncertain cases to symbolic reasoning.

\section{Conclusion and Future Work}
\label{sec:conclusion_future_work}

In this work, we proposed a hybrid neuro-symbolic framework for uncertainty-aware parking prediction in urban environments. Building on the strengths of Bayesian Neural Networks for probabilistic modelling and symbolic reasoning systems for interpretability and structure, we introduced two loosely coupled strategies to combine these paradigms: uncertainty-guided fallback and context-constrained refinement.

Through extensive evaluation on real-world parking sensor data under baseline, data scarcity, and noisy conditions, we demonstrated that our proposed methods improve predictive accuracy and robustness compared to standalone neural or symbolic models. In particular, the refinement-based strategy (Method 2) consistently outperformed all baselines, validating the benefit of integrating logical constraints into the prediction process. Although the observed improvements in accuracy are often within a 1–3\% range, they carry meaningful implications for real-world parking management systems, where even small gains can translate into reduced search times, improved guidance reliability, and more efficient traffic flow.

Moreover, Method 2 shows consistent performance across a range of data quality scenarios, demonstrating strong generalisability and resilience to uncertainty—key requirements for deployment in dynamic urban environments. Beyond accuracy, the hybrid neuro-symbolic framework enhances interoperability by enabling the model to incorporate domain-specific rules and adapt to new policy constraints or city-specific regulations without retraining the entire network. This flexibility, along with its robustness, positions our approach as a practical and scalable solution for intelligent parking prediction and similar context-aware forecasting tasks.

These results suggest that neuro-symbolic integration is a promising direction for building resilient and explainable AI systems for smart city infrastructure. By incorporating uncertainty-awareness and rule-based logic, such systems can better handle imperfect data and provide insights that are meaningful to human stakeholders.

Future work will explore tighter coupling strategies, including end-to-end differentiable neuro-symbolic models, as well as real-time deployment in multi-modal transportation environments. We also aim to extend this approach to broader context-aware prediction tasks where uncertainty and interpretability are critical.

\bibliographystyle{IEEEtran}
\bibliography{references/Bayesian_NNs,references/Neuro_Symbolic_AI,references/Parking_Prediction}
\vspace{12pt}

\end{document}